\documentclass[11pt,a4paper]{article}
\usepackage[hyperref]{eacl2021}
\usepackage{times}
\usepackage{latexsym}

\usepackage{microtype}

\usepackage{booktabs}
\usepackage[inline]{enumitem}

\usepackage{url}
\usepackage{graphicx}
\usepackage{pifont}
\usepackage{pgfplots}
\usepackage{tikz}
\usepackage{xspace}
\usepackage{comment}
\newcommand{\alert}[1]{\textcolor{red}{#1}}
\newcommand{\todo}[1]{}
\newcommand{\maybe}[1]{}

\newcommand{\nospacetext}[1]{\makebox[0pt][l]{#1}}

\newcommand{\pandora}{\textsc{Pandora}\xspace}

\title{\pandora Talks: Personality and Demographics on Reddit}
\author{Matej Gjurkovi{\'{c}} \quad Mladen Karan \quad Iva Vukojevi{\'{c}} \quad Mihaela Bo{\v{s}}njak \quad Jan {\v{S}}najder\\
Text Analysis and Knowledge Engineering Lab\\
Faculty of Electrical Engineering and Computing, University of Zagreb\\
Unska 3, 10000 Zagreb, Croatia \\
\tt name.surname@fer.hr
}

\date{}

\begin{document}
\pgfplotsset{width=7cm,compat=newest}
\maketitle
\begin{abstract}

Personality and demographics are important variables in social sciences and computational sociolinguistics.
However, datasets with both personality and demographic labels are scarce. To
address this, we present \pandora, the first \alert{\maybe{large-scale}} dataset of Reddit comments of 10k users partially labeled with three personality models and demographics (age, gender, and  location), including 1.6k users labeled with the well-established Big 5 personality model. We
showcase the usefulness of this dataset on three experiments, where we leverage
the more readily available data from other personality models to predict the
Big 5 traits, analyze gender classification biases arising from
psycho-demographic variables, and carry out a confirmatory and exploratory
analysis based on psychological theories. Finally, we present benchmark
prediction models for all personality and demographic variables.

\end{abstract}

\section{Introduction}

Personality and demographics describe differences between people at the
individual and group level. This makes them important for much of social
sciences research, where they may be used as either target or control
variables.  One field that can greatly benefit from textual datasets with
personality and demographic data is computational sociolinguistics
\cite{nguyen2016computational}, which uses NLP methods to study language use in
society. 
%Recently, the availability of large
%amounts of textual data on social media has attracted considerable interest in
%, obtained either manually or predicted automatically from text using 
%author profiling models \cite{daelemans2019overview,rangel2018overview}.
%\maybe{With the advent of big data,
%recent years have seen an increased interest in computational social science \cite{lazer2009computational}. 
% In particular,} 

%has found many practical applications in recommender systems, marketing, and
%conversational systems. Personality and demographic data of 

Conversely, personality and demographic data can be useful in the development of NLP systems. Recent advances in machine learning have brought significant improvements in NLP
systems' performance across many tasks, but these typically come
at the cost of more complex and less interpretable models, often susceptible to biases \cite{chang2019bias}. 
%focus of the community is now shifting toward explainability (e.g.,~ 
%\citet{serrano2019attention,pryzant2018deconfounded}).} 
%Interpetability is
%especially important in social sciences, where it relates to notions of
%validity and reliability \cite{miller2019explanation}. 
%
Biases are commonly caused by societal biases present in data, and eliminating
them requires a thorough understanding of the data used to train the model. One
way to do this is to consider 
%advantage of the body of knowledge amassed in social
%sciences, in particular sociology and psychology -- and in particular
demographic and personality variables, as language use and interpretation is
affected by both. Incorporating these variables into the design and analysis of
NLP models can help interpret model's decisions, avoid societal biases, and
control for confounders.

%our field is discovering, for example how personality and demographics interact
%with language and each with other. 

%\todo{make it more clear}

The demographic variables of age, gender, and location have been widely used in computational sociolinguistics \citep{bamman2014gender,peersman2011predicting,eisenstein2010latent}, while in NLP there is ample work on predicting these variables or using them in other NLP tasks. 
%
% These demographic variables are easy to understand even for researchers outside social sciences, the labels are easy to get on social media, and they are often easy for people to detect based on content or style. In some languages (e.g., Slavic), the information about gender is intrinsically encoded.  
%\todo{make it more clear: it's easy to detect someones gender, and in some languages it is obvious} 
%
%
In contrast, advances in text-based personality research are lagging
behind.  This can be traced to the fact that (1) personality-labeled datasets are
scarce and (2) personality labels are much harder to infer from text than
demographic variables such as age and gender.  In addition, the few existing datasets have
serious limitations: a small number of authors or comments, comments of limited
length, non-anonymity, or topic bias.  While most of
these limitations have been addressed by the recently published MBTI9k Reddit
dataset \citep{gjurkovic-snajder-2018-reddit}, this dataset still has two deficiencies.
Firstly, it uses the Myers-Briggs Type Indicator (MBTI) model 
\cite{myers1990introduction}, which -- while popular among the general
public and in business -- is discredited by most personality psychologists
\cite{barbuto1997critique}. The alternative is the well-known Five Factor Model (or Big 5)
\cite{mccrae1992an}, which, however, is less popular, and
thus labels for it are harder to obtain. Another deficiency of MBTI9k
is the lack of demographics, limiting model interpretability and use in sociolinguistics.

%Partially because demographics has more dominant linguistic clues, and the differences in distribution of particular demographics per specific personalities can lead to incorrect predictions and misleading analysis.

Our work seeks to address these problems by introducing a new dataset --
\emph{Personality ANd Demographics Of Reddit Authors} (\pandora) -- the first dataset from Reddit labeled with personality and demographic data.
\pandora comprises over 17M comments written by more than 10k Reddit users,
labeled with Big 5 and/or two other personality models (MBTI, Enneagram),
alongside age, gender, location, and language. In particular, Big 5 labels are available for more than 1.6k users, who jointly produced more than 3M comments. 

\pandora provides exciting opportunities for sociolinguistic research and
development of NLP models. In this paper we showcase its usefulness through
three experiments. In the first, inspired by work on domain adaptation and
multitask learning, we show how the MBTI and Enneagram labels can be used to
predict the labels from the well-established Big 5 model. We leverage
the fact that more data is available for MBTI and Enneagram, and exploit the
correlations between the traits of the different models and
their manifestations in text. In the second experiment we demonstrate how the
complete psycho-demographic profile can help in pinpointing biases in gender
classification.  
%The issue of bias is particularly important in case of Reddit,
%since Reddit was already used to train a number of NLP models.  
We show that a
gender classifier trained on a large Reddit dataset fails to predict
gender for users with certain combinations of personality traits more often
than for other users. Finally, the third experiment showcases the usefulness of
\pandora in social sciences: building on existing theories from psychology, we
perform a confirmatory and exploratory analysis between
propensity for philosophy and certain psycho-demographic variables.

We also report on baselines for personality and demographics prediction on
\pandora. We treat Big 5 and other personality and demographics variables as
targets for supervised machine learning, and evaluate a number of
benchmark models with different feature sets. We make \pandora
available\footnote{\href{https://psy.takelab.fer.hr}{https://psy.takelab.fer.hr}} 
for the research community, in the hope this will
stimulate further research.

\section{Background and Related Work}

\paragraph{Personality models and assessment.}

Myers-Briggs Type Indicator (MBTI; \citeauthor{myers1990introduction}, \citeyear{myers1990introduction}) and Five Factor Model (FFM; \citeauthor{mccrae1992an}, \citeyear{mccrae1992an}) are two most commonly used personality models. Myers-Briggs Type Indicator (MBTI) categorizes people in 16 personality types defined by four dichotomies: Introversion/Extraversion (way of gaining energy), Sensing/iNtuition (way of gathering information), Thinking/Feeling (way of making decisions), and Judging/Perceiving (preferences in interacting with others). The main criticism of MBTI focuses on low validity \cite{bess2002bimodal, mccrae1989reinterpreting}. 

Contrary to MBTI, FFM \cite{mccrae1992an} has a dimensional approach to
personality and describes people as somewhere on the continuum of five
personality traits (Big 5): Extraversion (outgoingness), Agreeableness (care
for social harmony), Conscientiousness (orderliness and self-discipline),
Neuroticism (tendency to experience distress), and Openness to Experience 
(appreciation for art and intellectual stimuli). Big 5 personality traits are generally assessed using inventories e.g., personality tests.\footnote{The usual inventories for assessing Big 5 are International Personality Item
Pool (IPIP; \citeauthor{goldberg2006the}, \citeyear{goldberg2006the}), Revised
NEO Personality Inventory (NEO-PI-R; \citeauthor{costa1991facet},
\citeyear{costa1991facet}), or Big 5 Inventory (BFI;
\citeauthor{john1991thebig}, \citeyear{john1991thebig}). Another common
inventory is HEXACO  \cite{lee2018psychometric}, which adds a sixth trait,
Honesty-Humility. \maybe{Correlations between the same traits in those four
inventories are positive and moderate (BFI and NEO-PI-R;
\citeauthor{schmitt2007the}, \citeyear{schmitt2007the}) to high
(\citeauthor{john:1999}, \citeyear{john:1999}, \citeauthor{gow2005goldbergsc},
\citeyear{gow2005goldbergsc}).}} Moreover,
personality has been shown to relate to some demographic variables, including gender
\cite{schmitt2008why}, age \cite{soto2011age}, and location
\cite{schmitt2007the}. Results show that females score higher than males in agreeableness, extraversion, conscientiousness, and neuroticism \cite{schmitt2008why}, and that expression of all five traits subtly changes during lifetime \cite{soto2011age}. There is also evidence of correlations between
MBTI and FFM \cite{furnham1996big, mccrae1989reinterpreting}. 

\paragraph{NLP and personality.}

Research on personality and language developed from early works on essays
\cite{pennebaker1999linguistic,argamon2005lexical,luyckx2008personae}, emails
\cite{oberlander2006language}, EAR devices \cite{mehl2001electronically}, and
blogs \cite{iacobelli2011large}, followed by early research on social networks \cite{quercia2011our,golbeck2011predicting}. 
In recent years, most research is done on Facebook \cite{schwartz2013personality,celli2013workshop,park2015automatic,tandera2017personality,kulkarni_latent_2018,xue2018deep}, Twitter \cite{plank2015personality,verhoeven2016twisty,tighe-cheng-2018-modeling,ramos-etal-2018-building,celli2018},
and Reddit \cite{gjurkovic-snajder-2018-reddit,wu2020author2vec}. 
Due to labeling cost and privacy concerns, it has become increasingly challenging to obtain personality datasets, especially large-scale dataset are virtually nonexistent \todo{matej: mypersonality}. \citet{wiegmann-etal-2019-celebrity} provide an overview of the datasets, some of which are not publicly available. 

%\maybe{
After MyPersonality dataset \cite{kosinski2015facebook} became unavailable to the research community, subsequent research had to rely on the few smaller datasets based on essays \cite{pennebaker1999linguistic}, personality forums,\footnote{\href{http://www.kaggle.com/datasnaek/mbti-type}{http://www.kaggle.com/datasnaek/mbti-type}} Twitter \cite{plank2015personality,verhoeven2016twisty}, and a small portion of the MyPersonality  dataset \cite{kosinski2013private} used in PAN workshops \cite{celli2013workshop,celli2014workshop,rangel2015overview}.
%}
%
% \cite{majumder,sun2018deep,xue2018deep}
%\maybe{

To the best of our knowledge, the only work that attempted to compare prediction models for both MBTI and Big 5 is that of \citet{celli2018}, carried out on Twitter data. However, they did not leverage the MBTI labels in the prediction of Big 5 traits, as their dataset contained no users labeled with both personality models.
%}

As most recent personality predictions models are based on deep learning \cite{majumder,xue2018deep,rissola_crestani2019,wu2020author2vec,lynn-etal-2020-hierarchical,vu-etal-2020-predicting,mehta2020recent,mehta2020bottom},  large-scale multi-labeled datasets such as \pandora can be used to develop new architectures and minimize the risk of models overfitting to spurious correlations.

\paragraph{User Factor Adaption.} \todo{matej: check references} Another important line of research that would
benefit from datasets like \pandora is debiasing based on demographic data
\cite{Liu_2017,zhang2018mitigating,pryzant-etal-2018-deconfounded,elazar2018adversarial,huang-paul-2019-neural}.
Current research is done on demographics, with the exception of the work of 
\citet{lynn-etal-2017-human}, who use personality traits, albeit predicted.
Different social media sites attract different types of users, and we expect more research of this kind on Reddit, especially
considering that Reddit is the source of data for many studies on mental
health
\cite{de2016discovering,yates-etal-2017-depression,sekulic-etal-2018-just,cohan-etal-2018-smhd,turcan-mckeown-2019-dreaddit,sekulic-strube-2019-adapting}.

%\cite{gkotsis2016language,shen2017detecting,,zirikly-etal-2019-clpsych,sekulic-strube-2019-adapting}.

%\cite{Liu_2017}
%\cite{zhang2018mitigating}
%\cite{pryzant-etal-2018-deconfounded}
%\cite{park-etal-2018-reducing}

%\cite{wiegmann-etal-2019-celebrity}
%\cite{argamon2005lexical}
%\cite{mairesse2007using}
%\cite{pennebaker1999linguistic}
%\cite{verhoeven2016twisty}
%\cite{luyckx2008personae}
%\cite{oberlander2006language}
%\footnote{\href{http://www.kaggle.com/datasnaek/mbti-type}{http://www.kaggle.com/datasnaek/mbti-type}}
%\cite{golbeck2011predicting}
%\cite{quercia2011our}
%\cite{park2015automatic}
%\cite{lynn-etal-2017-human}
%\cite{celli2013workshop,rangel2015overview}
%\newcite{plank2015personality}
%\cite{plank2015personality}. \newcite{verhoeven2016twisty}
%\cite{kulkarni_latent_2018}
%\cite{huang-paul-2019-neural}
%\cite{schwartz2013personality}
%\cite{shing-etal-2018-expert}
%\cite{elazar2018adversarial}
%By personality type (big5, mbti, ...)
%By other demographics data
%\cite{sekulic-etal-2018-just}
%age
%\cite{bamman2014gender,sap2014developing,ciot2013gender}

%\paragraph{multi-target demographics papers}
%Colorado and Lynn
%- we have real personality, they don't
%- we use similar region location model
%- we have much more text available (how we stand in terms of users)
%- do they have all labels for all users? we don't
%- what is their source? reddit different (anonymity, multiple topics, quality language)

\section{\pandora Dataset}
\label{sec:dataset}

Reddit is one of the most popular websites worldwide. Its users,  Redditors,
spend most of their online time on site and have more page views than users of
other websites. This, along with the fact that users are anonymous and that the
website is organized in more than a million different topics (subreddits), makes
Reddit suitable for various kinds of sociolinguistic studies. To compile their
MBTI9k Reddit dataset, \citet{gjurkovic-snajder-2018-reddit} used the Pushift Reddit dataset \cite{baumgartner2020} to retrieve the comments dating back to 2015. We adopt MBTI9k as the starting
point for \pandora. 
%What makes it even more appealing as a
%source of data is that the database of all its comments and posts from 2005 is
%available via the Google Big Query platform. 
%
%The relative ease of labelling
%convenient for our study stems from its concept of flairs -- short descriptions
%that users introduce themselves on a particular subreddit. 
%

\paragraph{Ethical Research Statement.}
We are following the ethics code for psychological research by which researchers may dispense with informed consent of each participant for archival research, for which disclosure of responses would not place participants at risk of criminal or civil liability, or damage their financial standing, employability, or reputation, and if confidentiality is protected. As per Reddit User Agreement, users agree not to disclose sensitive information of other users and they consent that their comments are publicly available and exposed through API to other services. The users may request to have their content removed, and we have taken this into account by removing such content; future requests will be treated in the same way and escalated to Reddit. Our study has been approved by an academic IRB.
%Other ethical considerations are discussed in the Appendix.

\subsection{MBTI and Enneagram Labels} 
\label{sec:mbti}

\citet{gjurkovic-snajder-2018-reddit} relied
on \emph{flairs} to extract the MBTI labels. Flairs are short
descriptions with which users introduce themselves on various subreddits, and
on MBTI-related subreddits they typically report on MBTI test results. Owing to
the fact that MBTI labels are easily identifiable, they used regular
expressions to obtain the labels from flairs (and
occasionally from comments). We use their labels for \pandora, but 
additionally manually label for Enneagram, which users also
typically report in their flairs. In total, 9,084 users reported their MBTI
type in the flair, and 793 additionally reported their Enneagram type. Table~\ref{tbl:big5_mean_std_distribution} shows the distribution of MBTI types and dimensions
(we omit Enneagram due to space constraints).
%  \todo {mihaela: prikazat tablice za mbti distrib}
%\begin{table}
%  \centering
%{\small
%    \begin{tabular}{lrlr}
%\toprule
%    MBTI Type & Users & MBTI Type & Users \\
%        \midrule
%    INTP      & 2833  & ISTJ      & 194 \\
%    INTJ      & 1841  & ENFJ      & 162 \\
%    INFP      & 1071  & ISFP      & 123 \\
%    INFJ      & 1051  & ISFJ      & 109 \\
%    ENTP      & 627   & ESTP      & 71  \\
%    ENFJ      & 616   & ESFP      & 51  \\
%    ISTP      & 408   & ESTJ      & 43  \\
%    ENTJ      & 319   & ESFJ      & 29  \\
%    \midrule
%        MBTI Dimension & Users & MBTI Dimension & Users \\
%        \midrule
%        Introverted & 7134 & Extroverted & 1920 \\
%        Intuitive & 8024 & Sensing & 1030 \\
%        Thinking & 5837 & Feeling & 3217 \\
%        Perceiving & 5302 & Judging & 3752 \\
%        
%    \bottomrule
%\end{tabular}\caption{\label{tbl:mbti_distribution} MBTI types and dimensions for 9,048 users \todo{if we %need more space we can cut distribution per MBTI types}}}
%\end{table}
 
% \begin{table}
%   \centering
%{\small
%    \begin{tabular}{lr}
%	    \toprule
%    Description   & Range   \\
%    \midrule
%    very\_low     & 0--10   \\
%    low           & 10--20  \\
%    low\_average  & 20--40  \\
%    average       & 40--60  \\
%    average\_high & 60--80  \\
%    high          & 80--90  \\
%    very\_high    & 90--100 \\
%    \bottomrule
%\end{tabular}\caption{\label{tbl:desc_map_distribution} Normalized descriptions mapping to %percentiles.}} \todo{move this inline}
%\end{table}

\subsection{Big 5 Labels}

Obtaining Big 5 labels turned out to be more challenging.  Unlike MBTI and
Enneagram tests, \mbox{Big 5} tests result in a score for each of the five traits. Moreover, the score format itself is not standardized, thus scores are reported in various formats and they are typically reported not in flairs but in comments replying to posts which mention a specific online test. Normalization of scores poses a series of
challenges. Firstly, different
web sites use different \maybe{personality tests and }inventories (e.g., \emph{HEXACO},
\emph{NEO PI-R}, \emph{Aspect-scale}), some of which are publicly available
while others are proprietary. 
%; Table \ref{tbl:Big5_tests_distribution} shows the distribution of tests and inventory in \pandora. 
The different tests use
different names for traits (e.g., emotional stability as the opposite of
neuroticism) or use abbreviations (e.g., \emph{OCEAN}, where \emph{O} stands
for openness, etc.). Secondly, test scores may be reported as either raw
scores, percentages, or percentiles.  Percentiles may be calculated based on
the distribution of users that took the test or on distribution of specific groups of offline test-takers (e.g., students), in the latter case commonly adjusted for age and gender. Moreover, scores
can be either numeric or descriptive, the former being in
different ranges (e.g., \emph{-100--100}, \emph{0--100}, \emph{1--5}) and the
latter being different for each test (e.g., descriptions \emph{typical} and \emph{average} may map to the same underlying score). On top
of this, users may decide to copy-paste the results, describe them in their own words
(e.g., \emph{rock-bottom} for low score) -- often misspelling the names of the
traits -- or combine both. Lastly, in some cases the results do not
even come from inventory-based assessments but from text-based personality
prediction services (e.g., \emph{Apply Magic Sauce}
%\footnote{\label{magic_sauce}https://applymagicsauce.com}
and \emph{Watson Personality}).

\paragraph{Extraction.} The fact that Big 5 scores are reported in full-text
comments rather than flairs and that their form is not standardized makes it
difficult to extract the scores fully automatically. Instead, we opted for a
semiautomatic approach as follows.  First, we retrieved candidate comments
containing three traits most likely to be spelled correctly
(\emph{agreeableness}, \emph{openness}, and \emph{extraversion}). For each
comment, we retrieved the corresponding post and determined what test it refers
to based on the link provided, if the link was present. We first discarded all
comment referring to text-based prediction services, and then used a set of
regular expressions specific to the report of each test to extract personality
scores from the comment.  Next, we manually verified all the extracted scores
and the associated comments to ensure that the comments indeed refer to a Big 5
test report and that the scores have been extracted correctly. For about 80\%
of reports the scores were extracted correctly, while for the remaining 20\% we
extracted the scores manually. This resulted in Big 5 scores for 1027 users,
reported from 12 different tests. Left out from this procedure were the
comments for which the test is unknown, as they were replying to posts without
a link to the test. To also extract scores from these reports, we trained a
test identification classifier on the reports of the 1,008 users, using
character n-grams as features, and reaching an F1-macro score of 81.4\% on
held-out test data. We use this classifier to identify the tests referred to in
the remaining comments and repeat the previous score extraction procedure. This
yielded scores for additional 600 users, for a total of 1,608 users. 
%Table
%\ref{tbl:Big5_tests_distribution} shows the distribution of
%personality tests and inventories.

% \begin{table}
% \centering
%{\small
%    \begin{tabular}{llrr}
%    \toprule
%    & &\multicolumn{2}{c}{\#\,Users}\\
%    \cmidrule(lr){3-4}
%%    Online test               & Used Inventory          & Gold & Pred\\
%    \midrule
%    Truity               & Proprietary
%     
%   % \alert{Goldberg's unipolar markers (\citeyear{goldberg1992development}) markers}  
%    & 378      & 319       \\
%
%    Understand Myself   & Big 5 Aspects       & 268      & 146     \\ 
%    IPIP 120             & IPIP-NEO      & 120      & 82      \\
%    IPIP 300             & IPIP-NEO       & 60       & 17      \\
%    Personality Assesor & BFI         & 66       & 10      \\
%    HEXACO               & HEXACO-PI-R               & 49       & 1       \\
%    Outofservice         & BFI               & 38       & 11      \\
%    Qualtrics            & --                   & 19       & 8       \\
%    123test              & IPIP-NEO       & 11       & 6       \\
%    BFI-2                 & BFI & 1 & 0 \\
%    See My Personality     & IPIP NEO& 1 & 0 \\
%    \bottomrule
%\end{tabular}\caption{\label{tbl:Big5_tests_distribution} User counts for gold and predicted online tests}}
%\end{table}

\paragraph{Normalization.} To normalize the extracted scores, we first heuristically mapped score
descriptions of various tests 
%
%to a unified set of descriptions (\emph{very\_low}, \emph{low},
%\emph{low\_average}, \emph{average}, \emph{average\_high}, \emph{high}, \emph{very\_high}), which
%we then converted 
to numeric values in the 0--100 range in increments of 10. As
mentioned, scores may refer to either raw scores, percentiles, or
descriptions. Both percentiles and raw scores are mostly reported on the same 0--100 scale, so we refer to the information on the test used to interpret the
score correctly. Finally, we convert raw scores and percentages reported by Truity\footnote{https://www.truity.com/} and HEXACO\footnote{http://hexaco.org/hexaco-online} to percentiles based on score distribution parameters. HEXACO reports distribution parameters publicly, while Truity provided us with parameters of the distribution of their test-takers. 

Finally, for all users labeled with Big 5 labels, we retrieved all their
comments from the year 2015 onward, and add these to the MBTI dataset from
\S\ref{sec:mbti}. The resulting dataset consists of 17,640,062 comments
written by 10,288 users. There are 393 users labeled with both Big 5 and MBTI.

\subsection{Demographic Labels} 
\label{ref:demographics}

To obtain age, gender, and location labels, we again turn to textual
descriptions provided in flairs.  For each of the 10,228 users, we collected
all the distinct flairs from all their comments in the dataset, and then
manually inspected these flairs for age, gender, and location information. For
users who reported their age in two or more flairs at different time points, we
consider the age from most recent one. Additionally, we extract comment-level self-reports of users' age (e.g., \emph{I'm 18 years old}) and gender (e.g., \emph{I'm female/male}).  As for
location, users report location at different levels, mostly countries, states,
and cities, but also continents and regions. We normalize location names, and
map countries to country codes, countries to continents, and states to
regions.
%\footnote{For mapping states to regions, there are different regional
%divisions for the U.S. and Canada. We used five regions for US, and three for
%Canada (for one region there was no users).}  
%Table~\ref{tbl:location_dist}
%shows that most users are from English speaking countries, and regionally evenly distributed in US and Canada. 
Most users are from English speaking countries, and regionally evenly distributed in US and Canada (cf.~Appendix). 
%Lastly, we use language
%identification \cite{joulin2016bag,joulin2016fasttext} at the comment level and
%calculate, for each user, the percentage of comments written in English. 
Table~\ref{tbl:big5_mean_std_distribution} shows the average number per user. Lastly,
Table~\ref{tbl:intersection} gives intersection counts between personality models and
other demographic variables.

\subsection{Analysis}

\begin{figure}
    \centering
    \vspace{-1em}
    \hspace*{-0.5cm}
    \includegraphics[scale=0.28]{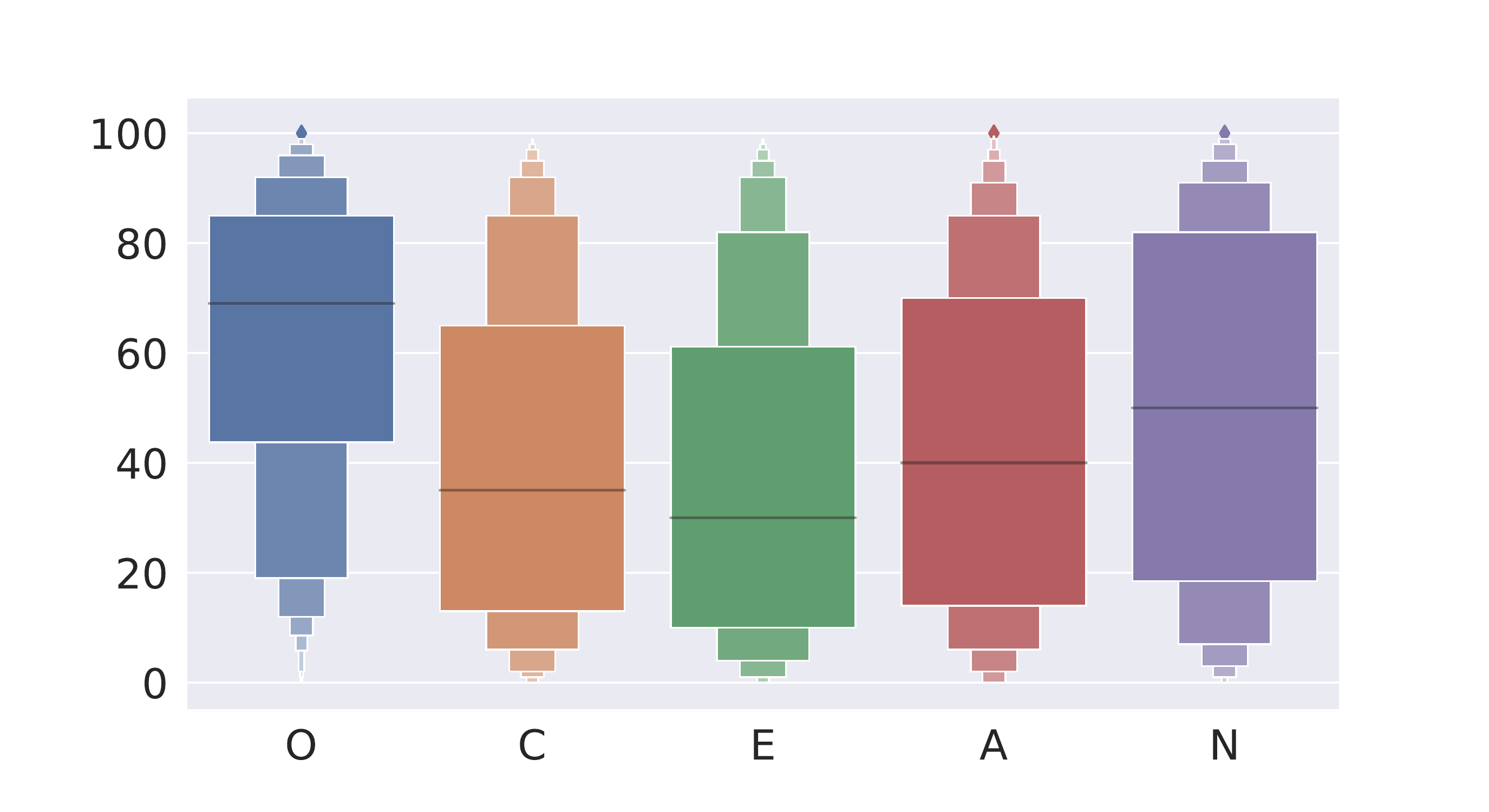}
    \vspace*{-0.5em}
    \caption{Distribution  of  Big  5  percentile  scores}
    \label{fig:distribBig5}
\end{figure}

Table~\ref{tbl:big5_mean_std_distribution} and Figure~\ref{fig:distribBig5} show the distributions of Big 5
scores per trait.\footnote{%
As noted by \citet{gjurkovic-snajder-2018-reddit}, due to various selection biases involved our dataset may not be representative of Reddit users, and it is certainly not representative of internet users or the general population.}
We observe that the
average user in our dataset is average on neuroticism, more open, and less
extraverted, agreeable, and
conscientious. Furthermore, males are on average younger, less agreeable, and neurotic than females. Similarly, Table~\ref{tbl:big5_mean_std_distribution} shows that MBTI
users have a preference for introversion, intuition (i.e., openness), thinking (i.e., less agreeable), and
perceiving (less conscientious). This is not surprising if we look at
Table~\ref{tbl:mbti_big5_corr_valid}, which shows high correlation between
particular MBTI dimensions or Enneagram types, and Big 5 traits. Correlations between Big 5 and MBTI follow the same pattern as correlations from existing psychological research \cite{mccrae1989reinterpreting}.

%Correlations between Enneagram and Big 5 are interpretable e.g. extraversion is positively correlated with Enneagram type 7 which describes an upbeat, ongoing person and negatively with type 5 which is more reserved and usually introverted. 

%Our next experiment is motivated by high correlations between personality models, the existence of users having two or more of them labeled, and the fact that there are more MBTI and Enneagram users then Big 5 labeled users.

  \begin{table}[t!]
   \centering
{\small
	\setlength{\tabcolsep}{5pt}
	\begin{tabular}{@{}lrrr@{}}

	    \toprule
%	    & \multicolumn{2}{c}{All} & \multicolumn{2}{c}{Male} & \multicolumn{2}{c}{Female}\\
    Big 5 Trait   & All   & Females & Males \\
    \midrule
    Openness           & 62.5  & 62.9 & 64.3\\
    Conscientiousness  & 40.2  & 43.3 & 41.6\\
    Extraversion       & 37.4  & 39.7 & 37.6\\
    Agreeableness      & 42.4  & 44.1 & 38.9\\
    Neuroticism        & 49.8  & 51.6 & 46.9\\
    \midrule 
    Age                & 25.7  & 26.7 & 25.6\\
    \midrule 
    \# Comments        & 1819  & 2004 & 3055\\ 
%    \# EN comments     & 1714  & 1908 & 2878\\
    
    \midrule
    \midrule
        MBTI Dimension & \#\,Users & MBTI Dimension & \#\,Users \\
        \midrule 
        Introverted & 7134 & Extraverted & 1920 \\
        Intuitive & 8024 & Sensing & 1030 \\
        Thinking & 5837 & Feeling & 3217 \\
        Perceiving & 5302 & Judging & 3752 \\
    \bottomrule
\end{tabular}\caption{\label{tbl:big5_mean_std_distribution}Means of Big 5 percentile scores (n=1608), age (n=2324), number of comments per user (n=10,255) and distribution of MBTI traits (n=9054)}}
\end{table}
 
%	is_female	0.0	1.0
%en_comments	mean	2878.610	1908.138
%std	5304.733	4851.693
%num_comments	mean	3055.431	2004.443
%std	5645.486	5204.545 
 
%%	is_female	0.0	1.0
%& 64.343 & 62.916 & 27.958 & 25.298
%& 41.567 & 43.306 & 31.337 & 30.650
%& 37.627 & 39.728 & 30.648 & 30.5 
%& 38.879 & 44.116 & 31.219 & 31.2
%& 46.874 & 51.574 & 31.772 & 31.4%  Openness           & 62.454 &  27.781 & 64.343 & 62.916 & 27.958 & 25.2985%  Conscientiousness  & .42&30.395  & 41.567 & 43.306 & 31.337 & 30.650   5%  Extroversion       & 37.3752 30.479  & 37.627 &8.728 &4.648 & 30.223 5%  Agreeableness     & 42.397 231.041  & 38.879 & 44.116 & 31.219431.009  5%  Neuroticism & .483 & 3274    & 46.874 & 51.574 & 31.772 & 3140        \\2%\todo{do we want lang.4 info at all?4\todo{ad8escriptive statistics about dataset: mean std .4all traits, age, per gender --8place 8le.4
% \todo{8 that we leave out some additional information about languages}

  \begin{table}
   \centering
{\small
    \begin{tabular}{lrrrr}
	    \toprule
       Variable &  Big 5  & MBTI  & Enneagram  &  Unique \\
    \midrule
    Gender & 599 & 2695 & 345 & 3084 \\
    Female & 232 & 1184 & 149 & 1331\\
    Male &  367 & 1511 & 196 & 1753\\
    Age & 638 & 1890 & 290 & 2324\\
    Country & 235 & 1984 & 182 & 2146\\
    Region & 74 & 800 & 65 & 852  \\
    Big 5 & -- & 393 & 64 &1608 \\
    MBTI & 393 & -- & 793 & 9084 \\
    Enneagram & 64 & 793 & -- & 794\\

    \bottomrule
\end{tabular}\caption{\label{tbl:intersection} Intersection details for personality models and the total number of unique labels}}
\end{table}

\begin{table}[t]

  \centering
{\small
%<<<<<<< HEAD
\begin{tabular}{lrrrrr}
\toprule 
%Gold & & & Predicted & \\
MBTI / Big 5 & O &C & E & A & N \\
\midrule
Introverted       & --.062  & --.062	 &\textbf{--.748}  & --.055 &\textbf{.157}	 \\
Intuitive         &\textbf{ .434}  & --.027 &--.042 & .030 &.065 \\
Thinking          &--.027	 & \textbf{.138}	 &--.043  & \textbf{--.554}&  \textbf{--.341}\\
Perceiving        &\textbf{.132} &\textbf{--.575}	 &\textbf{.145}  & .055& .031\\
\midrule
Enneagram 1 & --.139 & \textbf{.271}& --.012 & .004 & --.163 \\
Enneagram 2 &.038 & \textbf{.299}& .042 & \textbf{.278} & --.034 \\
Enneagram 3 & .188 & .004& .143 & --.069 & --.097 \\
Enneagram 4 & .087 & --.078& --.137 & \textbf{.320} &\textbf{.342} \\
Enneagram 5 & --.064 & .006 & \textbf{--.358} & --.157 & --.040 \\
Enneagram 6 & --.026 & .003 & --.053 & --.007 & \textbf{.276} \\
Enneagram 7 & .015 & \textbf{--.347} & \textbf{.393} & --.119 & \textbf{--.356} \\
Enneagram 8 & --.127 & .230& .234 & \textbf{--.363} & --.179 \\
Enneagram 9 & --.003 & --.155& --.028 & .018 & .090\\
\bottomrule
\end{tabular}\caption{\label{tbl:mbti_big5_corr_valid} Correlations between gold MBTI, Enneagram, and Big 5. Significant correlations (p\textless{.05}) are bolded.}
}
\end{table}

\section{Experiments}

Coupling linguistic data with psycho-demographic profiles sets the stage for
many interesting research questions. We showcase this with three experiments on
\pandora.

\subsection{Predicting Big 5 with MBTI/Enneagram}
\label{sec:mbtibig5}

MBTI and Enneagram are considerably more popular than Big 5 among the social
media users. This makes it relatively easy to obtain the MBTI and Enneagram
labels (\S\ref{sec:mbti}), and develop well-performing prediction models using
supervised machine learning.  On the other hand, validity of MBTI and Enneagram
has been severely criticized \cite{barbuto1997critique,thyer2015}, which is why
they are virtually not used in psychological research.
This experiment investigates whether we can combine the best of both worlds:
leverage the more abundant MBTI/Enneagram labels in \pandora to
predict Big 5 traits from text. We hypothesize that the questionable psychological
validity of MBTI/Enneagram labels 
can be compensated by their number. We base this on
moderate to strong correlations observed between the personality models
(Table~\ref{tbl:mbti_big5_corr_valid}) and the presence of a considerable
number of users with multiple labels (Table~\ref{tbl:intersection}).

We frame the experiment as a domain adaptation task of transferring 
MBTI/Enneagram labels to Big 5 labels, using a simple domain
adaptation approach from \citep{daume-iii-2007-frustratingly} (cf.~Appendix for more details). We train four text-based MBTI
classifiers on a subset of \pandora users for which we have MBTI labels but no Big 5 labels. We then apply these classifiers on a subset of \pandora users for which we have both MBTI and Big 5 labels, obtaining a type-level accuracy of MBTI prediction of 45\%. 
Table~\ref{tbl:mbti_big5_pred_corr_valid} shows correlations between MBTI and
Big 5 gold labels and predicted MBTI labels (cf.~Appendix for Enneagram correlations). As expected, we observe lower
overall correlations in comparison with correlations on gold labels
(Table~\ref{tbl:mbti_big5_corr_valid}). The main observable difference is that
extraversion is now moderately correlated with predicted MBTI intuitive dimension. 
As majority of Big 5 traits significantly correlate with more than one MBTI dimension, we use these scores as features for training five
regression models, one for each Big 5 trait. 
Lastly, we apply both classifiers on the subset of \pandora users for which we have Big 5 labels but no MBTI labels (serving as domain
adaptation target set). We use MBTI classifiers to obtain scores for the
four MBTI dimensions, and then feed these to Big 5 models to obtain
predictions for the five traits. The resulting correlations (Table~\ref{tbl:enne_mbti_big5_corr}) clearly
indicate that predictions based on MBTI 
help in predicting 
Big 5 traits.
Furthermore, the results justify the use of regression models as predicted Big 5 traits are more
correlated with gold Big 5 traits then predicted MBTI dimensions, with the exception of conscientiousness, which is significantly correlated with perceiving/judging MBTI dimension. For instance, predicted openness is a better predictor of openness than the intuitive dimension.

 \begin{table}[t!]

  \centering
{\small
\begin{tabular}{lrrrr}
\toprule 
  & \multicolumn{4}{c}{Predicted} \\
  \cmidrule(){2-5}
 %\cmidrule(c){2-5} 
%\cmidrule(l){1}\cmidrule(rrrr){2-5}
%Gold & & & Predicted & \\
Gold  & I/E & N/S & T/F & P/J \\
\midrule
O   & --.094 &\textbf{.251}  & --.087 & .088  \\
C & --.003 & .033  & .085  & \textbf{--.419} \\
E      &\textbf{--.516} & \textbf{.118}  & --.142 & --.002 \\
A     & .064  & .068  & \textbf{--.406} & .003  \\
N       & .076  & --.026 & \textbf{--.234} & .007  \\
\midrule
I/E       & \textbf{.513}  & --.096 & .023  & --.066 \\
N/S      & .046  & \textbf{.411}  & --.043 & .032  \\
T/F          & --.061 & --.036 & \textbf{.627}  & \textbf{.141}  \\
P/J        & \textbf{--.108} & --.033 & .083  & \textbf{.587} \\
\bottomrule
\end{tabular}\caption{\label{tbl:mbti_big5_pred_corr_valid} Correlations between predicted MBTI, Enneagram and Big 5 with gold Big 5 traits on users that reported both MBTI and Big 5. Significant correlations (p\textless{.05}) are shown in bold.}
}
\end{table}

% \usepackage{graphicx}
\begin{table}[]
  \centering
{\small
\begin{tabular}{crrrrr}
\toprule
Predicted & O & C & E & A & N \\
\midrule
I/E                       & \textbf{--.082}   & .039             & \textbf{--.262}       & --.003        & --.002      \\
N/S                       & \textbf{.127}    & --.021            & .049        & \textbf{.060}         & .001       \\
T/F                       & --.001   & .038             & --.039       & \textbf{--.259}        & \textbf{--.172}      \\
P/J                       & .018    & \textbf{--0.41}            & .007        & .034         & .039       \\
\midrule
O                       & \textbf{.147}    & \textbf{--.082}            & \textbf{.212}       & \textbf{.145}       & \textbf{.070}    \\
C                       & --.007   & \textbf{.237}            & .013        & \textbf{--.112}        & \textbf{--.090}     \\
E                       & .098    & --.028            & \textbf{.272}      & .044         & .022       \\
A                       & .006   & \textbf{--.079}           & .023        & \textbf{.264}         &\textbf{.176}       \\
N                       & --.048   & --.025            & --.042       &\textbf{.231}         & \textbf{.162}       \\
%\midrule
%Enn. 1                          & .002    & .032             & -.028       & .047         & .025       %\\
%Enn. 2                          & -.011   & \textbf{.108}             & .030        & \textbf{.135}    %     & \textbf{.046}       \\
%Enn. 3                          & \textbf{.085}   & .014             & \textbf{.071}        & -.064    %    & \textbf{-.069}      \\
%Enn. 4                          & -.041   & -.017            & -.033       & \textbf{.166}         & %\textbf{.159}       \\
%Enn. 5                          & \textbf{.067}    & -.035            & \textbf{-.060}       & %\textbf{-.121}        & \textbf{-.076}      \\
%Enn. 6                          & -.051   & .004             & -.035       & .046         & %\textbf{.113}       \\
%Enn. 7                          & -.043   & -.019            & \textbf{.078}        &\textbf{ -.085}   %     & \textbf{-.088}     \\
%Enn. 8                          & .022    & -.044            & \textbf{.063}        &\textbf{ -.129}   %     &\textbf{ -.075}      \\
%Enn. 9                         & -.034   & -0.016            & \textbf{-.102}       & .041         & %-.005     \\
\bottomrule
\end{tabular}\caption{\label{tbl:enne_mbti_big5_corr} Correlations between predicted MBTI, Enneagram and Big 5 with gold Big 5 traits on users that reported only Big 5 traits. Significant correlations (p\textless{.05}) are shown in bold.}
}
\end{table}

\subsection{Gender Classification Bias} 
\label{genderpred}

Gender classification from text is a fundamental task in author profiling, and
in particular author profiling on social media has recently received a lot of
attention from the NLP community
\cite{bamman2014gender,sap2014developing,ciot2013gender}.  
%\maybe{This is in part due
%to its usefulness in advertising and recommendation, but also because gender,
%together with age, is a variable many studies in computational sociolinguistics
%want to control for.} 
Additionally, gender is often in the spotlight of research
of fairness and bias in NLP \cite{sun-etal-2019-mitigating}. 
Biases are often introduced by demographic and other imbalances in training
data. Here we look at personality profile as a potential source of bias,
and set out to investigate whether a simple gender classification model trained on
Reddit exhibits biases that could be traced back to personality
traits. This is an important issue, given that Reddit is often used as a source
of data for training NLP models, e.g., 
\citep{zhang-etall-2017-char,cheng-etal-2017-factored,henderson-etal-2019-repository,sekulic-strube-2019-adapting}.

To build a gender classifier, we retrieve a separate Reddit dataset and label
it automatically for gender. To this end, we again rely on flairs, using
strings``/f/'' and ``/m/'' as female and male gender indicators,
respectively.\footnote{%
Although the construct of gender is not binary, we limit our present analysis to users who reported binary gender to obtain a more balanced dataset for bias analysis.}
This method yields a 98.5\% precision on
\pandora. From the 34k users that used these patterns in their flairs, we sampled a balanced dataset
of 24,954 users and retrieved over 30M of their comments, removing  
quoted text and all comments shorter than five words. Next,
we aggregate the comments per user, and divide the users in an 80\%--20\% train-test split. 
For classification, we use logistic regression with 500-dimensional SVD vectors derived from Tf-Idf word n-grams. The test accuracy of the classifier was 89.9\%.  The accuracy of the classifier on 3,084 users from \pandora with known gender was 89.3\%.

We now turn to bias analysis. On \pandora, the classifier failed to predict the
correct gender for 8.1\% male (142/1743) and 14.4\% female (192/1331) users. As
this is a statistically significant difference (p\textless0.05 with
two-proportion Z-Test), we conclude that the classifier is biased. To
investigate this further, we divide male and female users into those for which
the predictions were correct and those for which they were incorrect. We then test
for statistically significant differences (using two-proportion Z-test for
binary variables and Kruskal-Wallis H-test for continuous variables) of
psycho-demographic variables between correctly and incorrectly classified cases
for both groups. 
%\alert{The motivation behind this is to detect biases in the
%classifier with respect to such variables and differences in those biases for
%males and females.} 
Results are shown in Table~\ref{tbl:fm_predictors}.
Differences are statistically significant for thinking and perceiving MBTI
dimensions for both females and males, for extraversion Big 5 trait for males,
and for age in females.  Thinking and perceiving preference for females makes
them more likely to be misclassified for males, and the reverse holds for
males.  Furthermore, the gender of more extraverted males is more likely to be
misclassified. When it comes to age, younger females are more often in
misclassified group. These findings clearly indicate that a complete
psycho-demographic profile is a useful tool for bias analysis of machine
learning models trained on social media text.
\todo{jan: pristupi za kontrolu biasa na temelju r3}

%We also observe significant differences in total number of comments between the
%groups (males), number of English comments (males), and the percentage of
%English comments in all comments (females and males). We again turn to our
%dataset and find that the number of comments is significantly (p\textless0.05)
%and positively correlated with thinking (r=0.13 for males, 0.07 females) and
%age (r=0.12, 0.11) for both genders and negatively for intuition (-0.06) for
%females, which explains correlations between the number of comments and biased
%performance.

\begin{table}[]
  \centering
{\small

    \setlength{\tabcolsep}{4.9pt}
\begin{tabular}{@{}lrrrrrr@{}}

\toprule
& \multicolumn{3}{c}{Female} & \multicolumn{3}{c}{Male}\\
\cmidrule(lr){2-4}\cmidrule(lr){5-7}
Variable  & \ding{51} & \ding{56} & $\Delta$  & \ding{51} & \ding{56} & $\Delta$ \\
\midrule
Age                &  26.78    &  25.83  & \textbf{0.95\nospacetext{$^*$}}   & 25.46   & 26.90 & --1.44  \\
\midrule
I/E        & 0.78    & 0.72    & 0.06   & 0.76    & 0.82  & --0.06 \\
N/S            & 0.86    & 0.91    & --0.05   & 0.92    & 0.93 & --0.01 \\
T/F           &  0.47   &  0.64   & \textbf{--0.17\nospacetext{$^{***}$}}   &  0.61     &  0.29   &  \textbf{0.32\nospacetext{$^{***}$}} \\
P/J         &  0.39    &  0.56  &\textbf{--0.17\nospacetext{$^{***}$}}   &  0.53     &  0.39  & \textbf{0.14\nospacetext{$^{***}$}} \\
\midrule
O           & 61.40   & 68.18 & --6,78   & 64.11   & 67.20 & --3.09 \\
C         & 45.28   & 36.44 & 8.84  & 41.10   & 47.50 & --6.40 \\
E       & 40.67   & 36.44 & 4.23  &  36.68    &  49.60  &\textbf{--12.92\nospacetext{$^*$}}\\
A      & 45.07   & 40.78 & 4.29  & 38.43   & 44.70  & --6.27\\
N        & 50.95   & 53.72 & --2.77 & 46.81   & 47.50  & --0.69\\
%\midrule
%\#\,com      & 1899 & 2626 & -727 &  3145  &  2034  & \textbf{1112\nospacetext{***}}\\
%\#\,en\_com       & 1824 & 2403 & -579 &  2964  &  1900  & \textbf{1064\nospacetext{***}} \\
%\%\,en\_com      &  96.17    &  94.43  & \textbf{1.74\nospacetext{**}}   &  94.51    &  95.75  & \textbf{-1.24\nospacetext{**}} \\
%\midrule
%Gender   & 0.88    & 0.29 & 0.59    & 0.12    & 0.69  & -0.57 \\
\bottomrule
\end{tabular}\caption{\label{tbl:fm_predictors} Differences in means of psycho-demographic variables per gender and classification outcome. Significant correlations (*p\textless{.05}, {***}p\textless{.001}) are in bold.}
}
\end{table}

%#increased awarness of gender fairness. biased classificators
%we motivaate the need of complete psychodemographics profile to show that even there is fair classficator if we %look only at gender target, it is not so if take complete profile

% \begin{table}
%  \centering
%{\small
%\begin{tabular}{lrr}
%%\toprule
% & Male\_pred & Female\_pred \\
% \midrule 
% Male & 1611 & 142 \\
% Female & 192 & 1139 \\
% \bottomrule
%\end{tabular}\caption{\label{tbl:fm_confusion} Gender classification confusion matrix on psycho dataset. }}
%\end{table}

%\paragraph{Language, location, and demographics.}

\subsection{Propensity for Philosophy}

Our last experiment investigates the usefulness of \pandora for research in
social sciences. One obvious type of use cases are confirmatory studies which
aim to replicate present theories and findings on a dataset that has been
obtained in a manner different from typical datasets in the field. 
Another type
of use cases are 
%studies that 
%investigate how these theories are manifested in
%language of online talk, as well as 
exploratory studies that seek to identify
new relations between psycho-demographic variables manifested in online talk. Here
we present a use case of both types. We focus on propensity for philosophy of Reddit users 
(manifested as propensity for philosophical topics in online discussions), and
seek to confirm its hypothesized positive relationship with openness to
experiences \cite{johnson2014measuring, dollinger1996on}, cognitive processing
(e.g., insight), and readability index. We expect this to be confirmed since
all four variables share proneness to higher intellectual engagement. For exploratory analysis, we extend our analysis to emotion variables.

We conducted the analysis using hierarchical regression analysis with propensity
for philosophical topics as the criterion variable and demographics,
personality, emotions, cognitive processing, and text readability as
predictors. As a measure of propensity for philosophical topics, we compute
the ``philosophy'' feature (frequency of philosophical words) from Empath
\cite{fast2016empath} for each  user's comments. Similarly,
for the predictors we compute posemo, negemo, and insight features from LIWC \cite{pennebaker2015development} and Flesh-Kincaid Grade Level (F-K GL)
readability score \cite{kincaid1975derivation}.\footnote{We counted the frequencies per comment, divided it by total number of words in a comment, multiplied with 100, and averaged for total comments.} Emotion variables are inserted for the exploratory analysis.  
In the hierarchical regression analysis, demographics were added as control variables in the first step, Big 5 traits were added in the second step, emotion variables in the third step, and finally insight feature as a cognitive inclination variable and F-K GL readability index were added in the last step. The sample comprises 430 Reddit users, 273 males and 157 females, with the mean age of 26.79 (SD=7.954), who all had gold labels of gender, age, and Big 5.

\begin{table}
  \centering
{\small
    \setlength{\tabcolsep}{4.9pt}
    \begin{tabular}{lrrrrr}
	    \toprule
	    &\multicolumn{5}{c}{Regression coefficients}\\
	    \cmidrule(lr){2-6}
Predictors      & Step 1	& Step 2 &	Step 3 &	Step 4 &	Step 5\\
    \midrule
Gender &	--.26\nospacetext{$^{**}$} &	--.24\nospacetext{$^{**}$} &	--.20\nospacetext{$^{**}$} &	--.19\nospacetext{$^{**}$} &	--.17\nospacetext{$^{**}$}\\
Age &	--.01 &	--.03 &	--.02 &	.00 &	.01\\
O &	-- &	.20\nospacetext{$^{**}$} &	.19\nospacetext{$^{**}$} &	.15\nospacetext{$^{**}$} &	 .10\nospacetext{$^{**}$}\\
C &	-- &	.01 & 	.05 & 	.08 & 	.07\\
E & 	-- &	.02 &	.03 & 	.04 & 	.04\\
A &	-- &	--.12\nospacetext{$^*$} &	--.05 &	--.05 &	--.06\\
N &	-- &	--.04 &	--.03 &	.01 &	.02\\
posemo &	-- &	-- &	.15\nospacetext{$^{**}$} &	.17\nospacetext{$^{**}$} &	.03\\
negemo &	-- &	-- &	.29\nospacetext{$^{**}$} &	.27\nospacetext{$^{**}$} &	.29\nospacetext{$^{**}$}\\
insight &	-- &	-- &	-- &	.36\nospacetext{$^{**}$} &	.27\nospacetext{$^{**}$}\\
F-K GL  &	-- &	-- &	-- &	-- &	.34\nospacetext{$^{**}$}\\
    \midrule
$R^{2}$ &	.07 & 	.12 & 	.22 &	.34 & 	.43\\
Adjusted $R^{2}$ &	.06 & 	.11 & 	.20 &	 .32 &	.41\\  
$R^{2}$ change &	.07\nospacetext{$^{**}$} &	 .06\nospacetext{$^{**}$} &	.10\nospacetext{$^{**}$} &	 .12\nospacetext{$^{**}$} &	.09\nospacetext{$^{**}$}\\
\bottomrule
\end{tabular}\caption{\label{tbl:philosophy} Hierarchical regression of propensity for philosophical topics (``philosophy'' feature from Empath) on gender, age, Big 5 personality traits, Flesh-Kincaid Grade Level readability scores, positive and negative emotions features, and insight feature as predictors (n=430). The table shows regression coefficients and the goodness of fit as measured by $R^2$, adjusted $R^2$, and $R^2$ change. Significant correlations: *p\textless{.05}, **p\textless{.01},  ***p\textless{.001}}.}
\end{table}

The analysis yields interesting results.%
\footnote{Multivariate normality and multicollinearity were satisfied, and homoscedasticity was satisfied after removing 14 outliers based on standardized residuals.} Firstly, as much as the 41\% of variance in the ``philosophy'' feature is explained by the 11 predictors. Secondly, openness to experiences, readability index, and insight are, as expected, all significant and positive predictors of the ``philosophy'' feature. Agreeableness was a negative significant predictor before adding the emotion variables. This is not surprising, as people low in agreeableness are less likely to pander to others, and agreeableness shows significant correlations with both positive (.20) and negative emotions (-.13). Thirdly, the results imply alluring associations with emotion variables. Negative emotions were clearly positive predictors of frequency of discussing philosophical topics. However, positive emotions were a significant predictor until the last step when F-K GL was added to the model. This was due to moderate correlation between posemo and F-K GL (-0.40). Lastly, males had higher frequency of words related to philosophy than females. To sum up, the hypothesis is confirmed and exploratory analysis yields interesting results which could motivate further research.

\section{Prediction Models}

In this section we describe baseline models for predicting personality and
demographic variables from user comments in \pandora.

We consider the following sets of features: (1) \textbf{N-grams:} Tf-Idf weighted
 1--3 word ngrams and 2--5 character n-grams; (2)  \textbf{Stylistic}: the
counts of words, characters, and syllables, mono/polysyllable words, long
words, unique words, as well as all readability metrics implemented in
Textacy\footnote{\href{https://chartbeat-labs.github.io/textacy}{https://chartbeat-labs.github.io/textacy}};
(3) \textbf{Dictionaries}: words mapped to Tf-Idf categories from LIWC
\cite{pennebaker2015development}, Empath \cite{fast2016empath}, and NRC Emotion
Lexicon \cite{mohammad2013crowdsourcing} dictionaries; (4) \textbf{Gender}: predictions of
the gender classifier from \S\ref{genderpred}; (5) \textbf{Subreddit
distributions}: a matrix where each row is a distribution of post
counts across all subreddits for a particular user, reduced using PCA to 50
features per user; (6) \textbf{Subreddit other}: counts of downs, score,
gilded, ups, as well as the controversiality scores for a comment; (7)
\textbf{Named entities}: the number of named entities per comment, as extracted
using Spacy;\footnote{\href{https://spacy.io/}{https://spacy.io/}} (8)
\textbf{Part-of-speech}: counts for each part-of-speech; (9)
\textbf{Predictions} (only for predicting Big 5 traits): MBTI/Enneagram
predictions obtained by a classifier built on held-out data.
Features (2), (4), and (6--9) are calculated at the level of individual comments and
aggregated to min, max, mean, standard deviation, and median values 
for each user.
%The exceptions to this are the  Subreddit PCA features, which
%are by construction at the user level, and the N-gram and Dictionaries features
%which are calculated by considering the concatenation of comments for each user
%as a single text and applying tf-idf weighting.

We build six regression models (age and Big 5 traits) and
eight classification models (four MBTI dimensions, gender, region,
Enneagram). We experiment with linear/logistic regression (LR)
%and ExtraTrees (ET) models
from sklearn \cite{pedregosa2011scikit} and deep learning models (NN).
We trained a separate NN model for each task. In each model, a single user is represented as a matrix, with rows representing the user's comments. The comments were encoded using 1024-dimensional vectors derived using BERT \cite{devlin-etal-2019-bert}. BERT comment vectors are fed into convolution layers, max pooling, and several fully connected layers. Hyperparameters and additional information can be found in the Appendix.% \ref{sec:appendix}.
% Limited by the available computational resources, we used the most recent 100 comments of each user. 
%The models consist of three parts: a convolutional layer, a max-pooling layer, and several fully connected (FC) layers. 
%Convolutional kernels are as wide as BERT’s representation and slide vertically over the matrix to aggregate information from several comments. We tried different kernel sizes varying from 2 to 6, and different numbers of kernels $M$ varying from 4 to 6. Outputs of the convolutional layer are first sliced into a fixed number of $K$ slices and then subject to max pooling. This results in $M$ vectors of length $K$ per user, one for each kernel, which are passed to several FC layers with Leaky ReLU activations. Regularization (L2-norm and dropout) is applied only to FC layers.

%Evaluation is done via 5-fold cross-validation, while performing a separate stratified split
%for each target. We use regression F-tests to select top-$K$ features, where in
%each fold the hyperparameters of the models and $K$ are optimized via grid
%search on a held-out data. 

We evaluate the models using 5-fold cross-validation with a separate stratified split for each target.
We use regression F-tests to select top-$K$ features, and
optimize model hyperparameters and $K$ on held-out data for each fold separately.

%The models are
%evaluated with macro-F1 and Pearson correlation coefficient, respectively.

Results are shown in Table~\ref{tbl:baselineres}. LR performs best when using
only the n-gram features. 
%(ET does benefit from other features, but reaches a
%comparable performance; we omit these results due to space limits.)
An exception are Big 5 trait predictions, which benefit considerably from adding
the MBTI/Enneagram predictions as features, building on Section~\ref{sec:mbtibig5} and Table~\ref{tbl:mbti_big5_corr_valid}. 
Also, using 1000 comments rather than last 100 (as in NN) increased scores up to 5 points.
% We additionally trained the models for
% different number of comments per user. Results show that more comments (up to 1000, compared to 100)
% increase scores by up to 5 points, compared to training only on last 100
% comments per user. 
%\todo{TODO vidjet jel sa 2000 jednako dobro tj. jel plafon vec dosegnut.}

\begin{table}[]

  \centering
{\small
\setlength{\tabcolsep}{4.5pt}
\begin{tabular}{@{}lcccccc@{}}
\toprule 
  & \multicolumn{5}{c}{LR} & \\
  \cmidrule(lr){2-6}
  & NO & N & O & NOP & NP & NN \\
\midrule
\multicolumn{7}{c}{Classification (Macro-averaged F1 score)} \\
\midrule
Introverted &        .649 &        \textbf{.654} &        .559 & -- & -- & .546\\
Intuitive   &        .599 &        \textbf{.606} &        .518 & -- & -- & .528\\
Thinking     &       .730 &        \textbf{.739} &        .678 & -- & -- & .634\\
Perceiving    &      .626 &        \textbf{.642} &        .586 & -- & -- & .566\\
Enneagram &          .155 &        \textbf{.251} &        .145 & -- & -- & .143\\
Gender         &     .889 &        \textbf{.904} &        .825 & -- & -- & .843\\
Region          &    .206 &        \textbf{.626} &        .144 & -- & -- & .478\\
\midrule
\multicolumn{7}{c}{Regression (Pearson correlation coefficient)} \\
\midrule
Agreeableness  &     .181 &        .232 &        .085 &  .237 &         \textbf{.270} & .210\\ 
Openness        &         .235 &        \textbf{.265} &        .180 &    .235 &         .250 & .159\\
Conscientiousness &        .194 &        .162 &        .093 &   .245 &         \textbf{.273} & .120\\  
Neuroticism        &       .194 &        .244 &        .138 &   .266 &         \textbf{.283} & .149\\ 
Extraversion &              .271 &        .327 &        .058 &  .286 &         \textbf{.387} & .167\\ 
Age &        .704 &        \textbf{.750} &        .469 & -- & -- & .396\\
\bottomrule
\end{tabular}\caption{\label{tbl:baselineres}Prediction results for the different traits for LR and NN models. For the LR model, we show the results for different feature combinations, including N-grams (N), MBTI/Enneagram predictions (P), and all other features (O). %Scores are macro-averaged F1 score for classification tasks and Pearson correlation coefficient for regression tasks.%
Best results are shown in bold.}
}
\end{table}

\section{Conclusion}

\pandora\ dataset comprises 17M comments,
personality, and demographic labels for over 10k Reddit users, including 1.6k users with Big 5 labels. To our knowledge,
this is the first Reddit dataset with Big 5 traits, and also
the first covering multiple personality models (Big 5, MBTI, Enneagram).  We
showcased the usefulness of \pandora with three experiments, showing (1)
how more readily available MBTI/Enneagram labels can be used to estimate Big 5
traits, (2) that a gender classifier trained on Reddit exhibits bias on users
of certain personality traits, and (3) that certain psycho-demographic
variables are good predictors of propensity for philosophy of Reddit users.  We
also trained and evaluated benchmark prediction models for all psycho-demographic
variables. The poor performance of deep learning baseline models, the rich set of labels, and the large number of comments per user in \pandora suggest that further efforts should be directed toward efficient user representations and more advanced deep learning architectures.%, such as multi-task and adversarial models. 

\section*{Acknowledgements}

We thank the reviewers for their remarks. This work has been fully supported by the Croatian Science Foundation under the project IP-2020-02-8671 PSYTXT (``Computational Models for Text-Based Personality Prediction and Analysis'').

\bibliography{eacl2021} 
\bibliographystyle{acl_natbib} 

\clearpage 

\appendix

\section{Demographics of \pandora}
\label{sec:appendix}

\begin{table}[h]
  \centering
{\small
    \begin{tabular}{lr|lr}
	    \toprule
	        Continent      & \#\,Users & Continent    & \#\,Users \\
    \midrule
   North America & 1299  & Africa        & 4     \\
   Europe        & 580   & South America & 24    \\
    Asia          & 103  & Oceania       & 85    \\
    \midrule
    Country        & \#\,Users  & Region            & \#\,Users \\
    \midrule
    US             & 1107       & US West           & 208       \\
    Canada         & 180        & US Midwest        & 153       \\
    UK             & 164        & US Southeast      & 144       \\
    Australia      & 72         & US Northeast      & 138       \\
    Germany        & 53         & US Southwest      & 100       \\
    Netherlands    & 37         & Canada West       & 50        \\
    Sweden         & 33         & Canada East       & 44        \\
    \bottomrule
\end{tabular}\caption{\label{tbl:location_dist} 
Geographical distribution of users per continent, country, and region (for US and Canada)}}
\end{table}

 \begin{table}[h]
 \centering
{\small
    \begin{tabular}{lr}
	    \toprule
    Language & \#\,Comments \\
    \midrule
English         & 16637211 \\
Spanish         & 87309    \\
French          & 72651    \\
Italian         & 64819    \\
German          & 63492    \\
Portuguese      & 32037    \\
Dutch           & 30219    \\
Esperanto       & 19501    \\
Swedish         & 16880    \\
Polish          & 15134    \\
    \bottomrule
\end{tabular}\caption{\label{tbl:lang_dist} 
Language distribution}}
\end{table}

Table~\ref{tbl:location_dist}
shows that most users are from English speaking countries, and regionally evenly distributed in US and Canada.  
For mapping states to regions, there are different regional
divisions for the U.S. and Canada. We used five regions for US, and three for Canada (for one region there was no users). 

Additionally, for each comment we ran fast text based language identification.\footnote{https://fasttext.cc/docs/en/language-identification.html} Table~\ref{tbl:lang_dist} shows the number of comments for top 10 languages.

\section{Additional Information on Personality Scores}

Table~\ref{tbl:mbti_distribution} shows counts of all 16 MBTI types. Four MBTI types (\emph{INTP}, \emph{INTJ}, \emph{INFP} and \emph{INFJ}) account for 75 percent of all users. This indicates that there is a shift in personality distributions in contrast to the general public.
Table~\ref{tbl:big5_results_distribution} contains means and standard deviations for descriptions and percentiles of every Big 5 trait.  Table~\ref{tbl:Big5_tests_distribution} shows the distribution of tests and their inventories in \pandora.  

\begin{table}
 \centering
{\small
    \begin{tabular}{lrlr}
\toprule
    MBTI Type & Users & MBTI Type & Users \\
        \midrule
    INTP      & 2833  & ISTJ      & 194 \\
    INTJ      & 1841  & ENFJ      & 162 \\
    INFP      & 1071  & ISFP      & 123 \\
    INFJ      & 1051  & ISFJ      & 109 \\
    ENTP      & 627   & ESTP      & 71  \\
    ENFJ      & 616   & ESFP      & 51  \\
    ISTP      & 408   & ESTJ      & 43  \\
    ENTJ      & 319   & ESFJ      & 29  \\
    \bottomrule
\end{tabular}\caption{\label{tbl:mbti_distribution} 
MBTI types for 9,048 users 
}}
\end{table}
 
\begin{table}
 \centering
{\small
    \begin{tabular}{lll}
\toprule
    Trait             & Descriptions        & Percentiles \\%& Scores & ~ \\
    \midrule
    Agreeableness     & $50.10 \pm 29.10$      & $42.39 \pm 30.89$        \\%& 58.10  & ~ \\
    % ~                 & std  & 29.10        & 30.89       \\%& 21.96  & ~ \\
    % \midrule%\\
    Openness          & $67.37 \pm 26.76$        & $67.27 \pm 26.87$       \\%& 76.91  & ~ \\
    % ~                 & std  & 26.76        & 26.87       \\%& 15.14  & ~ \\
    % \midrule%\\
    Conscientiousness & $41.29 \pm  27.97$       & $40.48 \pm  30.22$      \\%& 50.67  & ~ \\
    % ~                 & std  & 27.97        & 30.22       \\%& 21.65  & ~ \\
    % \midrule%\\
    Extraversion      & $38.70 \pm  27.53$       & $37.09 \pm  31.16$      \\%& 43.15  & ~ \\ 
    % ~                 & std  & 27.53        & 31.16       \\%& 24.18  & ~ \\
    % \midrule%\\
    Neuroticism       & $55.95 \pm 31.11$        & $52.82 \pm  31.97$      \\%& 51.56  & ~ \\
    % ~                 & std  & 31.11        & 31.97       \\%& 24.70  & ~ \\
    \bottomrule
\end{tabular}\caption{\label{tbl:big5_results_distribution} Big 5 results distribution on different reported scales for 1,652 users}
%}\todo{switch to number$\pm$std, one row per trait}}
}
\end{table}

 \begin{table}
  \centering
{\small

\begin{tabular}{@{}llrr@{}}
    \toprule
    Online test                 & Based on inventory          & \#\,Users & \#\,Pred\\
    \midrule
    Truity               & IPIP, NEO-PI-R, \\
    &  Goldberg's (\citeyear{goldberg1992development}) \\ & markers         & 378      & 362       \\
    Understand\\Myself   & Big 5 Aspects       & 268      & 167     \\
    IPIP 120             & IPIP-NEO-120     & 120      & 83      \\
    IPIP 300             & IPIP-NEO-300       & 60       & 18      \\
    Personality\\Assesor & BFI         & 66       & 10      \\
    HEXACO               & HEXACO-PI-R               & 49       & 1       \\
    Outofservice         & BFI               & 38       & 11      \\
    Qualtrics            & --                   & 19       & 8       \\
    123test              & IPIP-NEO, NEO-PI-R       & 16       & 6       \\
    \bottomrule
\end{tabular}\caption{\label{tbl:Big5_tests_distribution} Big 5 personality test distribution in reports}
%\todo{if "inventory" is an established name for a "personality model", let's use that in the introduction, iva: i suggest changing here test- online test, inventory - based on inventory, Big 5 - BFI}
}

\end{table}

 \begin{table}
  \centering
{\small

\begin{tabular}{crrrrr}
\toprule
Predicted & O & C & E & A & N \\
\midrule
Enneagram 1         & .002              & .032              & --.028            & .047                    & .025       \\
Enneagram 2         & --.011            & \textbf{.108}     & .030              & \textbf{.135}           & \textbf{.046}       \\
Enneagram 3         & \textbf{.085}     & .014              & \textbf{.071}     & --.064                  & \textbf{--.069}      \\
Enneagram 4         & --.041            & --.017            & --.033            & \textbf{.166}           & \textbf{.159}       \\
Enneagram 5         & \textbf{.067}     & --.035             & \textbf{--.060}   & \textbf{--.121}          & \textbf{--.076}      \\
Enneagram 6         & --.051            & .004              & --.035             & .046                    & \textbf{.113}       \\
Enneagram 7         & --.043            & --.019            & \textbf{.078}     &\textbf{ --.085}         & \textbf{--.088}     \\
Enneagram 8         & .022              & --.044            & \textbf{.063}     &\textbf{ --.129}         &\textbf{--.075}      \\
Enneagram 9         & --.034             & --.016           & \textbf{--.102}   & .041                    & --.005     \\
\bottomrule
\end{tabular}\caption{\label{tbl:enne_corr} Correlations between Enneagram types and Big 5 traits. Significant correlations (p\textless{.05}) are shown in bold.}
}
\end{table}

\section{Predicting Big 5 with MBTI/Enneagram}

Here we describe in more details the setup for predicting Big 5 labels using MBTI/Enneagram labels.

We frame the experiment as a domain adaptation task of transferring 
MBTI/Enneagram labels to Big 5 labels, and use one of the simplest domain
adaptation approaches where we use source classifier (MBTI) predictions as features and linearly interpolate them on development set containing both MBTI and Big 5 to make predictions on Big 5 target set (e.g., \emph{PRED} and \emph{LININT} baselines from \citep{daume-iii-2007-frustratingly}). We first partition \pandora into three subsets: comments of users for
which we have both MBTI and Big 5 labels (M+B+, n=382), comments of users for
which we have the MBTI but no Big 5 labels (\mbox{M+B-}, n=8,691), and comments
of users for which we have the Big 5 but no MBTI labels (M-B+, n=1,588).
We then proceed in three steps.  In the first step, we train on M+B- four text-based MBTI
classifiers, one for each MBTI dimension (logistic regression, optimized with 5-fold CV, using 7000
filter-selected, Tf-Idf-weighed 1--5 word and character n-grams as features).

In the second step, we use text-based MBTI classifiers to obtain MBTI labels
on M+B+ (serving as domain adaptation source set), observing a type-level accuracy of 45\% (82.4\% for one-off prediction).
The
classifiers output probabilities, which can be interpreted as a score of the
corresponding MBTI dimension. As majority of Big 5 traits significantly correlate with more than one MBTI dimension, we use these scores as features for training five
regression models, one for each Big 5 trait (Ridge regression optimized with
5-fold CV). Additionally, we preformed a correlation analysis between Enneagram types and Big 5 traits. Results are shown in Table~\ref{tbl:enne_corr}.

In the third step, we apply both classifiers on M-B+ (serving as domain
adaptation target set): we first use MBTI classifiers to obtain scores for the
four MBTI dimensions, and then feed these to Big 5 regression models to obtain
predictions for the five traits. 

\section{Parameters of the DL Model}

The models consist of three parts: a convolutional layer, a max-pooling layer, and several fully connected (FC) layers. 
Convolutional kernels are as wide as BERT’s representation and slide vertically over the matrix to aggregate information from several comments. We tried different kernel sizes varying from 2 to 6, and different numbers of kernels $M$ varying from 4 to 6. Outputs of the convolutional layer are first sliced into a fixed number of $K$ slices and then subject to max pooling. This results in $M$ vectors of length $K$ per user, one for each kernel, which are passed to several FC layers with Leaky ReLU activations. Regularization (L2-norm and dropout) is applied only to FC layers.

\section{Learning Curves for the Logistic Regression Models}

\begin{figure}
    \centering
    %\vspace{-1em}
    %\hspace*{-0.5cm}
    \includegraphics[scale=0.54]{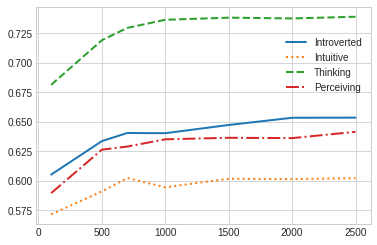}
    %\vspace*{-1.5em}
    \caption{Learning curves of logistic regression for MBTI trait prediction}
    \label{fig:lcmbti}
\end{figure}

\begin{figure}
    \centering
    %\vspace{-1em}
    %\hspace*{-0.5cm}
    \includegraphics[scale=0.54]{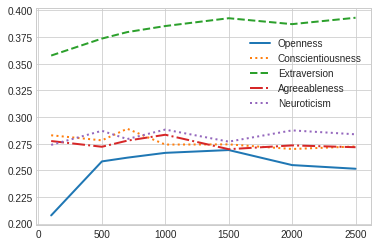}
    %\vspace*{-1.5em}
    \caption{Learning curves of logistic regression for Big 5 trait prediction}
    \label{fig:lcbig5}
\end{figure}

Figures~\ref{fig:lcmbti} and \ref{fig:lcbig5} show the learning curves for logistic regression model with 1-gram features: x-axis is the number of comments and y-axis is model's F1-macro score. Performance plateaus at around 1000 comments, showing little significant changes when increasing the number of comments used for training beyond that amount. 

\end{document}